\documentclass[runningheads]{llncs}
\usepackage{graphicx}
\usepackage{comment}
\usepackage{amsmath,amssymb} 
\usepackage{color}
\usepackage{algorithm}
\usepackage{algpseudocode}
\usepackage{tabulary,multirow,overpic,xcolor}
\usepackage{makecell}

\def\etal{\emph{et al}.}

\newcolumntype{C}[1]{>{\centering}m{#1}}

\begin{document}
	\pagestyle{headings}
	\mainmatter
	\def\ECCVSubNumber{}  
	
	\title{SF-Net: Single-Frame Supervision for \\Temporal Action Localization} 
	
	
	\titlerunning{SF-Net}
	%
	\author{Fan Ma\inst{1} \and
		Linchao Zhu\inst{1} \and
		Yi Yang\inst{1} \and
		Shengxin Zha\inst{2} \and
		Gourab Kundu\inst{2} \and \\
		Matt Feiszli\inst{2} \and
		Zheng Shou\inst{2}}
	\authorrunning{F. Ma et al.}
	%
	\institute{University of Technology Sydney, Australia \and
		Facebook}
	\maketitle
	
	\begin{abstract}
		{
			In this paper, we study an intermediate form of supervision, i.e., \textbf{single-frame supervision}, for temporal action localization (TAL).
			To obtain the single-frame supervision, the annotators are asked to identify only a single frame  \textit{within} the temporal window of an action. This can significantly reduce the labor cost of obtaining full supervision which requires annotating the action \textit{boundary}. Compared to the weak supervision that only annotates the video-level label, the single-frame supervision introduces extra temporal action signals while maintaining low annotation overhead.
			To make full use of such single-frame supervision, we propose a unified system called \textbf{SF-Net}.
			First, we propose to predict an actionness score for each video frame.
			Along with a typical category score, the actionness score can provide comprehensive information about the occurrence of a potential action and aid the temporal boundary refinement during inference.
			Second, we mine pseudo action and background frames based on the single-frame annotations.
			We identify pseudo action frames by adaptively expanding each annotated single frame to its nearby, contextual frames and we mine pseudo background frames from all the unannotated frames across multiple videos.
			Together with the ground-truth labeled frames, these pseudo-labeled frames are further used for training the classifier.
			In extensive experiments on THUMOS14, GTEA, and BEOID, SF-Net significantly improves upon state-of-the-art weakly-supervised methods in terms of both segment localization and single-frame localization.
			Notably, SF-Net achieves comparable results to its fully-supervised counterpart which requires much more resource intensive annotations. The code is available at \url{https://github.com/Flowerfan/SF-Net}.
		}
	\end{abstract}

	\section{Introduction}

	Recently, weakly-supervised Temporal Action Localization (TAL) has attracted substantial interest.  Given a training set containing only video-level labels, we aim to detect and classify each action instance in long, untrimmed testing videos.
	In the fully-supervised annotation, the annotators usually need to rollback the video for repeated watching to give the precise temporal boundary of an action instance when they notice an action while watching the video~\cite{zhao2019hacs}. For the weakly-supervised annotation, annotators just need to watch the video once to give labels. They can record the action class once they notice an unseen action.
	This significantly reduces annotation resources: video-level labels use fewer resources than annotating the start and end times in the fully-supervised setting.

	\begin{figure}[!t]
		\centering
		\includegraphics[width=\linewidth]{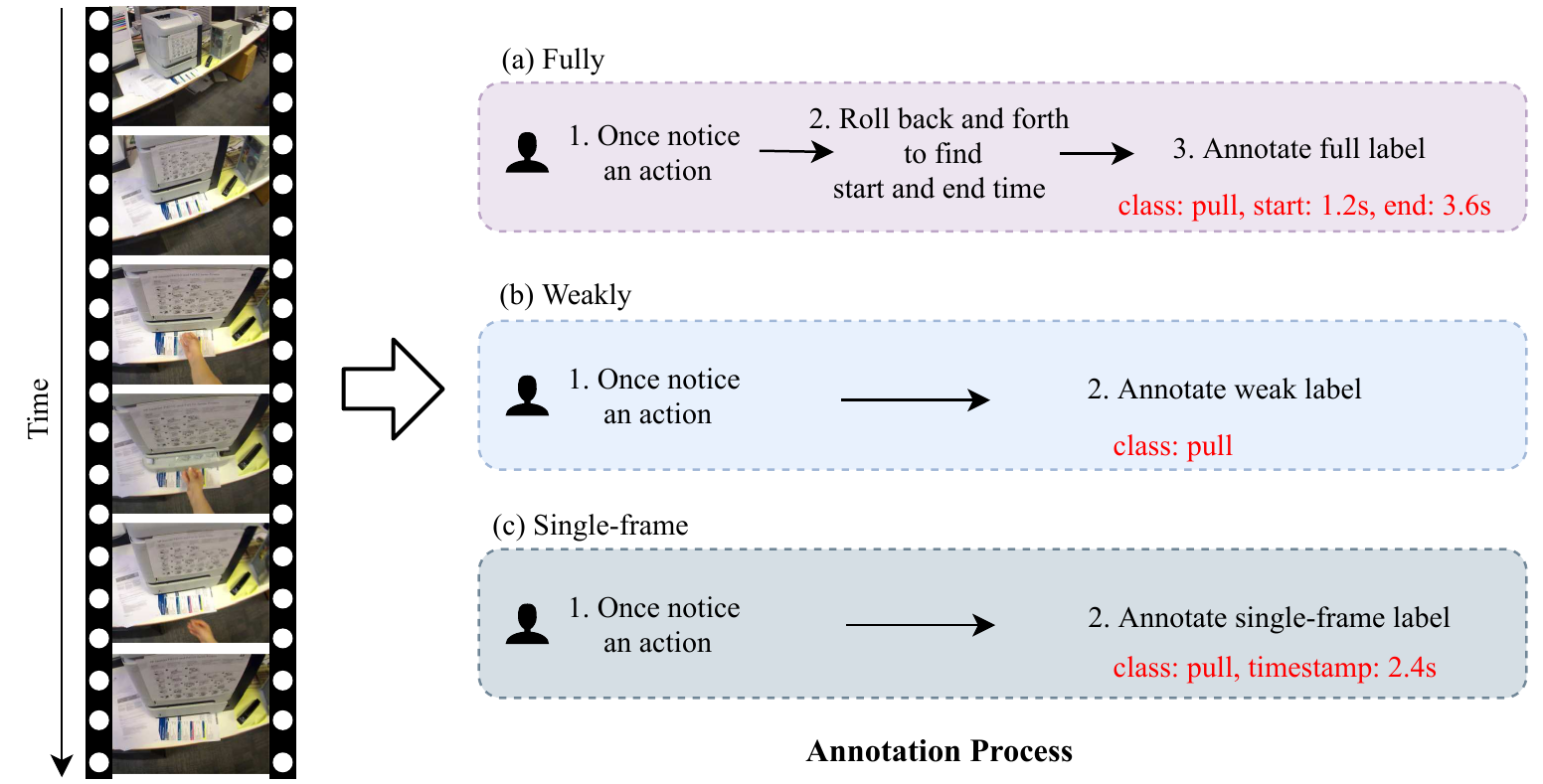}
		\caption{Different ways of annotating actions while watching a video. (a) Annotating actions in the fully-supervised way. The start and end time of each action instance are required to be annotated. (b) Annotating actions in the weakly-supervised setting. Only action classes are required to be given. (c) Annotating actions in our single-frame supervision. Each action instance should have one timestamp. Note that the time is automatically generated by the annotation tool. Compared to the weakly-supervised annotation, the single-frame annotation requires only a few extra pauses to annotate repeated seen actions in one video.}
		\label{fig:introduction}
	\end{figure}

	Despite the promising results achieved by state-of-the-art weakly-supervised TAL work \cite{nguyen2018weakly,paul2018w,shou2018autoloc}, their localization performance is still inferior to fully-supervised TAL work \cite{chao2018rethinking,lin2018bsn,Shou_2017_CVPR}.
	In order to bridge this gap, we are motivated to utilize single-frame supervision \cite{moltisanti2019action}: for each action instance, only one single positive frame is pointed out.
	The annotation process for single-frame supervision is almost the same as it in the weakly-supervised annotation. 
	The annotators only watch the video once to record the action class and timestamp when they notice each action. It significantly reduces annotation resources compared to full supervision.
	
	In the image domain, Bearman et al. \cite{bearman2016s} were the first to propose point supervision for image semantic segmentation. Annotating at point-level was extended by Mettes et al. \cite{mettes2016spot} to video domain for spatio-temporal localization, where each action frame requires one spatial point annotation during training. 
	Moltisanti et al. \cite{moltisanti2019action} further reduced the required resources by proposing single-frame supervision and developed a method that selects frames with a very high confidence score as pseudo action frames.
	
	However, \cite{moltisanti2019action} was designed for whole video classification. In order to make full use of single-frame supervision for our TAL task, the unique challenge of \textbf{localizing temporal boundaries of actions} remains unresolved.
	To address this challenge in TAL, we make three innovations to improve the localization model's capability in distinguishing background frames and action frames.
	First, we predict ``actionness'' at each frame which indicates the probability of being any actions. 
	Second, based on the actionness, we investigate a novel background mining algorithm to determine frames that are likely to be the background and leverage these pseudo background frames as additional supervision.
	Third, when labeling pseudo action frames, besides the frames with high confidence scores, we aim to determine more pseudo action frames and thus propose an action frame expansion algorithm.
	
	In addition, for many real-world applications, detecting precise start time and end time is overkill. Consider a reporter who wants to find some car accident shots in an archive of street camera videos: it is sufficient to retrieve a single frame for each accident, and the reporter can easily truncate clips of desired lengths.
	Thus, in addition to evaluating traditional segment localization in TAL, we also propose a new task called single-frame localization, which requires only localizing one frame per action instance.
	
	In summary, our contributions are three-fold:
	
	(1) 
	To our best knowledge, this is the first work to use single-frame supervision for the challenging problem of localizing temporal boundaries of actions. 
	We show that the single-frame annotation significantly saves annotation time compared to fully-supervised annotation.
	
	
	(2) We find that single-frame supervision can provide strong cue about the background. Thus, from frames that are not annotated, we propose two novel methods to mine likely background frames and action frames, respectively. These likely background and action timestamps are further used as pseudo ground truth for training.
	
	(3) We conduct extensive experiments on three benchmarks, and the performances on both segment localization and single-frame localization tasks are largely boosted.
	
	\section{Related Work}

	\textbf{Action recognition.} Action recognition has recently witnessed an increased focus on trimmed videos. Both temporal and spatial information is significant for classifying the video.
	Early works mainly employed hand-crafted features to solve this task. IDT~\cite{WangIDT} had been widely used across many video-related tasks. 
	Recently, various deep neural networks were proposed to encode spatial-temporal video information. 
	Two-stream network~\cite{simonyan2014two} adopted optical flow to learn temporal motion, which had been used in many latter works~\cite{CarreiraI3D,simonyan2014two,TSN2016ECCV}. 
	Many 3D convolutional networks~\cite{CarreiraI3D,Tran_2015_ICCV,Feichtenhofer_2019_ICCV} are also designed to learn action embeddings. Beyond fully-supervised action recognition, a few works focus on self-supervised video feature learning \cite{zhu2020actbert} and few-shot action recognition \cite{zhu2020lim}. In this paper, we focus on single-frame supervision for temporal action localization.
	
	\noindent\textbf{Point supervision.} Bearman et al. \cite{bearman2016s} first utilized the point supervision for image semantic segmentation. Mettes et al. \cite{mettes2016spot} extended it to spatio-temporal localization in video, where the action is pointed out by one spatial location in each action frame.
	We believe this is overkill for temporal localization, and demonstrate that single-frame supervision can achieve very promising results already. Recently, single-frame supervision has been used in \cite{moltisanti2019action} for video-level classification, but this work does not address identifying temporal boundaries.
	Note that Alwassel et al. \cite{alwassel2018action} proposed to spot action in the video during inference time but targeted detecting one action instance per class in one video while our proposed single-frame localization task aims to detect every instance in one video.

	\noindent\textbf{Fully-supervised temporal action localization.}
	Approaches of temporal action localization trained in full supervision have mainly followed a proposal-classification paradigm~\cite{chao2018rethinking,dai2017temporal,gao2017cascaded,lin2018bsn,Shou_2016_CVPR,Shou_2017_CVPR}, where temporal proposals are generated first and then classified. 
	Other categories of methods, including sequential decision-making~\cite{alwassel2018action}  and single-shot detectors~\cite{lin2017single} have also been studied.
	Given full temporal boundary annotations, the proposal-classification methods usually filter out the background frames at the proposal stage via a binary actionness classifier. 
	Activity completeness has also been studied in the temporal action localization task.
	Zhao et al.~\cite{zhao2017temporal} used a structural temporal pyramid pooling followed by an explicit binary classifier to evaluate the completeness of an action instance. 
	Yuan et al.~\cite{yuan2017temporal} structured an action into three parts to model its temporal evolution.
	Ch{\'e}ron~\etal~\cite{cheron2018flexible} handled the spatio-temporal action localization with various supervisions.
	Long et al.~\cite{long2019gaussian} proposed a Gaussian kernel to dynamically optimize temporal scale of action proposals. 
	However, these methods use fully temporal annotations, which are resource intensive.

	\noindent\textbf{Weakly-supervised temporal action localization.} 
	Multiple Instance Learning~(MIL) has been widely used in weakly-supervised temporal action localization. 
	Without temporal boundary annotations, temporal action score sequence has been widely used to generate action proposals~\cite{wang2017untrimmednets,nguyen2018weakly,Liu_2019_CVPR,Narayan_2019_ICCV}.
	Wang et al.~\cite{wang2017untrimmednets} proposed UntrimmedNet composed of a classification module and a selection module to reason about the temporal duration of action instances.
	Nguyen~\etal~\cite{nguyen2018weakly} introduced a sparsity regularization for video-level classification.
	Shou~\etal~\cite{shou2018autoloc} and Liu~\cite{liu2019weakly} investigated score contrast in the temporal dimension.
	Hide-and-Seek~\cite{singh2017hide} randomly removed frame sequences during training to force the network to respond to multiple relevant parts.
	Liu~\etal~\cite{Liu_2019_CVPR} proposed a multi-branch network to model the completeness of actions.
	Narayan~\etal~\cite{Narayan_2019_ICCV} introduced three-loss forms to guide the learning discriminative action features with enhanced localization capabilities.
	Nguyen~\cite{Nguyen_2019_ICCV} used attention modules to detect foreground and background for detecting actions.
	Despite the improvements over time, the performances of weakly-supervised methods are still inferior to the fully-supervised method.

	\section{Method}
	In this section, we define our tasks, present architecture of our SF-Net, and finally discuss details of training and inference, respectively.

	\subsection{Problem Definition}
	A training video can contain multiple action classes and multiple action instances.
	Unlike the full supervision setting, which provides temporal boundary annotation of each action instance, in our single-frame supervision setting, each instance only has one frame pointed out by annotator with timestamp $t$ and action class $y$. 
	Note that $y \in \{1,\dots, N_{c}\}$ where $N_{c}$ is the total number of classes and we use index 0 to represent the background class.
	
	Given a testing video, we perform two temporal localization tasks: (1) \textbf{Segment localization}. We detect the start time and end time for each action instance with its action class prediction.
	(2) \textbf{Single-frame localization}. We output the timestamp of each detected action instance with its action class prediction.
	The evaluation metrics for these two tasks are explained in Sec. 4.

	\begin{figure}[!t]
		\centering
		\includegraphics[width=\linewidth]{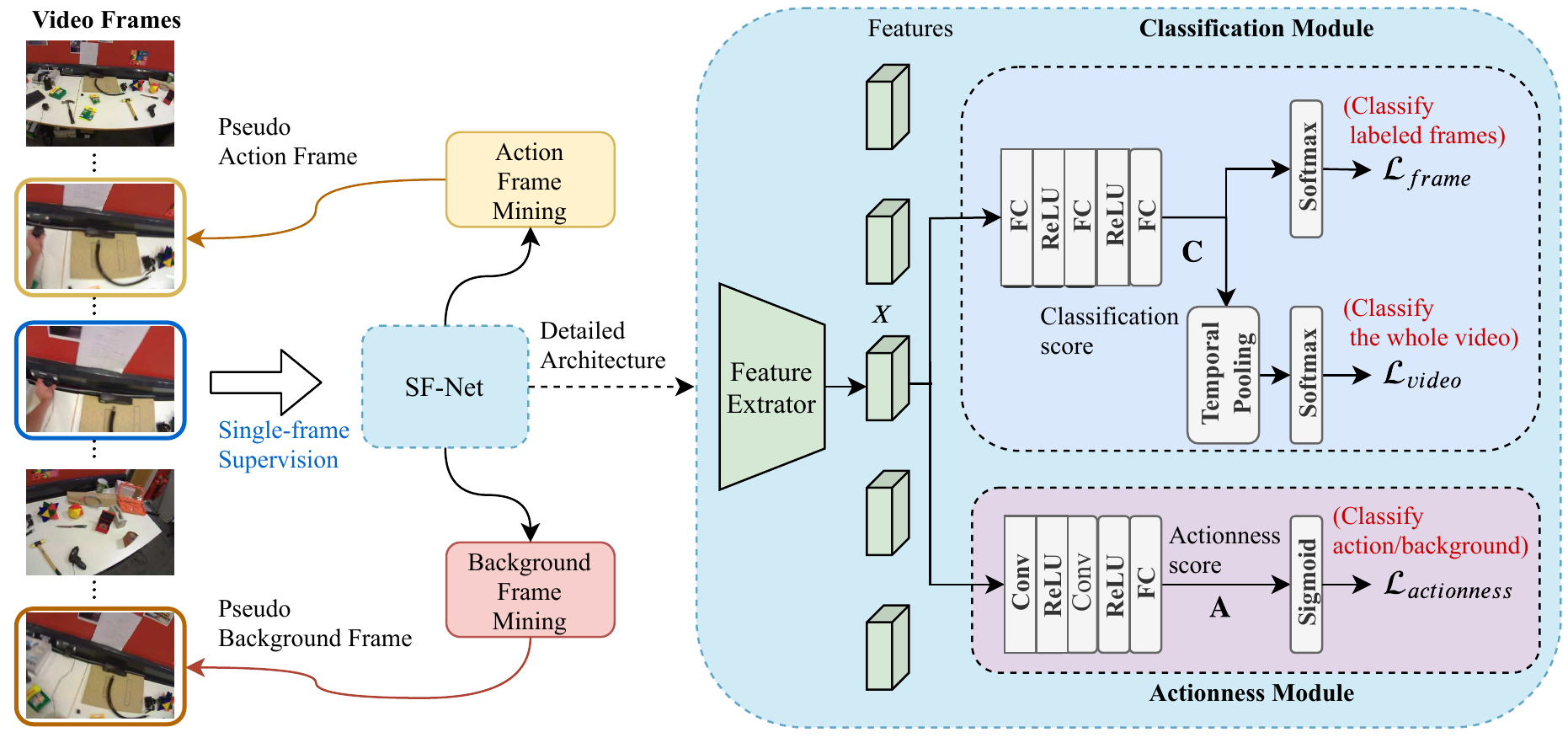}
		\caption{Overall training framework of our proposed SF-Net.
			Given single-frame supervision, we employ two novel frame mining strategies to label pseudo action frames and background frames.
			The detailed architecture of SF-Net is shown on the right.
			SF-Net consists of a classification module to classify each labeled frame and the whole video, and an actionness module to predict the probability of each frame being action.
			The classification module and actionness module are trained jointly with three losses explained in Sec. 3.3.
		}
		\label{fig:framework}
	\end{figure}

	\subsection{Framework}
	
	\textbf{Overview}. Our overall framework is presented in Fig.~\ref{fig:framework}. During training, learning from single-frame supervision, SF-Net mines pseudo action and background frames. Based on the labeled frames, we employ three losses to jointly train a classification module to classify each labeled frame and the whole video, and an actionness module to predict the probability of each frame being action. In the following, we outline the framework while details of frame mining strategies and different losses are described in Sec. 3.3.
	
	\noindent\textbf{Feature extraction}. For a training batch of N videos, the features of all frames are extracted and stored in a feature tensor  $X \in R^{N\times T \times D}$, where $D$ is the feature dimension, and $T$ is the number of frames.
	As different videos vary in the temporal length, we simply pad zeros when the number of frames in a video is less than $T$. 
	
	\noindent\textbf{Classification module}. 
	The classification module outputs the score of being each action class for all frames in the input video. To classify each labeled frame, we feed $X$ into three Fully-Connected~(FC) layers to get the classification score $C \in \mathcal{R}^{N\times T \times N_{c}+1}$. The classification score $C$ is then used to compute frame classification loss $\mathcal{L}_{frame}$.
	We also pool $C$ temporally as described in \cite{Narayan_2019_ICCV} to compute video-level classification loss $\mathcal{L}_{video}$.

	\noindent\textbf{Actionness module}. As shown in Fig.~\ref{fig:framework}, our model has an actionness branch of identifying positive action frames. Different from the classification module, the actionness module only produces a scalar for each frame to denote the probability of being contained in an action segment. To predict an actionness score, we feed $X$ into two temporal convolutional layers followed by one FC layer, resulting in an actionness score matrix $A \in \mathcal{R}^{N\times T}$. We apply sigmoid on $A$ and then compute a binary classification loss $\mathcal{L}_{actionness}$.

	\subsection{Pseudo Label Mining and Training Objectives}
	
	\subsubsection{Action classification at labeled frames.}
	We use cross entropy loss for the action frame classification.
	As there are $NT$ frames in the input batch of videos and most of the frames are unlabeled, we first filter the labeled frames for classification.
	Suppose we have $K$ labeled frames where $K\ll NT$.
	We can get classification activations of K labeled frames from $C$. These scores are fed to a Softmax layer to get classification probability $\mathbf{p}^{l} \in \mathcal{R}^{K\times N_{c}+1}$ for all labeled frames.  The classification loss of annotated frames in the batch of videos is formulated as:
	\begin{equation}
	\mathcal{L}_{frame}^{l} = -\frac{1}{K} \sum_{i}^{K} \mathbf{y}_{i} \text{log} \mathbf{p}^{l}_{i},
	\end{equation}
	where the $\mathbf{p}^{l}_{i}$ denote the prediction for the $i^{th}$ labeled action frame.
	
	\subsubsection{Pseudo labeling of frames.}
	
	With only a single label per action instance, the total number of positive examples is quite small and may be difficult to learn from.  While we do not use full temporal annotation, it is clear that actions are longer events spanning consecutive frames.
	To increase the temporal information available to the model, we design an action frame mining and a background frame mining strategy to introduce more frames into the training process.

	(a) Action frame mining:
	We treat each labeled action frame as an anchor frame for each action instance.  
	We first set the expand radius $r$ to limit the maximum expansion distance to the anchor frame at $t$.
	Then we expand the past from $t-1$ frame and the future from $t+1$ frame, separately. 
	Suppose the action class of the anchor frame is represented by $y_{i}$.
	If the current expanding frame has the same predicted label with the anchor frame, and the classification score at $y_{i}$ class is higher than that score of the anchor frame multiplying a predefined value $\xi$, we then annotate this frame with label $y_{i}$ and put it into the training pool. 
	Otherwise, we stop the expansion process for the current anchor frame.

	(b) Background frame mining:
	The background frames are also important and widely used in localization methods~\cite{Liu_2019_CVPR,Nguyen_2019_ICCV} to boost the model performance.  
	Since there is no background label under the single-frame supervision, our proposed model manages to localize background frames from all the unlabeled frames in the $N$ videos. 
	At the beginning, we do not have supervision about where the background frames are. But explicitly introducing a background class can avoid forcing classifying a frame into one of the action classes. Our proposed background frame mining algorithm can offer us the supervision needed for training such a background class so as to improve the discriminability of the classifier.
	Suppose we try to mine $\eta K$ background frames, we first gather the classification scores of all unlabeled frames from $C$. The $\eta$ is the ratio of background frames to labeled frames.
	These scores are then sorted along background class to select the top $\eta K$ scores $\mathbf{p}^{b} \in \mathcal{R}^{\eta K}$ as the score vector of the background frames. 
	The pseudo background classification loss is calculated on the top $\eta K$ frames by,
	\begin{equation}
	\mathcal{L}_{frame}^{b} = -\frac{1}{\eta K} \sum \text{log}~\mathbf{p}^{b},
	\end{equation}
	The background frame classification loss assists the model with identifying irrelevant frames. Different from background mining in~\cite{Liu_2019_CVPR,Nguyen_2019_ICCV} which either require extra computation source to generate background frames or adopt a complicated loss for optimization, we mining background frames across multiple videos and use the classification loss for optimization. 
	The selected pseudo background frames may have some noises in the initial training rounds. As the training evolves and the classifier’s discriminability improves, we are able to reduce the noises and detect background frames more correctly. With the more correct background frames as supervision signals, the classifier’s discriminability can be further boosted.
	In our experiments, we observed that this simple background mining strategy allows for better action localization results.
	We incorporate the background classification loss with the labeled frame classification loss to formulate the single-frame classification loss
	\begin{equation}
	\mathcal{L}_{frame} = \mathcal{L}_{frame}^{l}  + \frac{1}{N_{c}}\mathcal{L}_{frame}^{b}
	\end{equation}
	where $N_{c}$ is the number of action classes to leverage the influence from background class. 
	
	\begin{figure}[!t]
		\centering
		\includegraphics[width=0.7\linewidth]{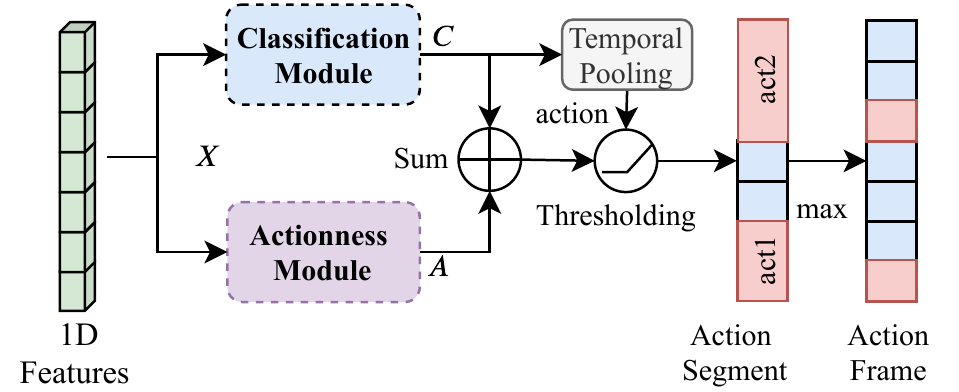}
		\caption{The inference framework of SF-Net. The classification module outputs the classification score $C$ of each frame for identifying possible target actions in the given video.
			The action module produces the actionness score determining the possibility of a frame containing target actions. The actionness score together with the classification score are used to generate action segment based on the threshold. }
		\label{fig:fig2}
	\end{figure}

	\subsubsection{Actionness prediction at labeled frames.}
	In fully-supervised TAL, various methods learn to generate action proposals that may contain the potential activities~\cite{xu2017r,lin2018bsn,chao2018rethinking}. 
	Motivated by this, we design the actionness module to make the model focus on frames relevant to target actions.
	Instead of producing the temporal segment in proposal methods, our actionness module produces the actionness score for each frame.
	The actionness module is in parallel with the classification module in our SF-Net. It offers extra information for temporal action localization.
	We first gather the actionness score $A^{l} \in \mathcal{R}^{K}$ of labeled frames in the training videos.
	The higher the value for a frame, the higher probability of that frame belongs to a target action. 
	We also use the background frame mining strategy to get the actionness score $A^{b} \in \mathcal{R}^{\eta K}$.
	The actionness loss is calculated by, 	
	\begin{equation}
	\mathcal{L}_{actionness} = -\frac{1}{K} \sum \log \sigma (A^{l}) - \frac{1}{\eta K} \sum \log(1 - \sigma(A^{b})),
	\end{equation}
	where $\sigma$ is the sigmoid function to scale the actionness score to $[0, 1]$.

	\subsubsection{Full objective.}
	
	We employ video-level loss as described in~\cite{Narayan_2019_ICCV} to tackle the problem of multi-label action classification at video-level. 
	For the $i^{th}$ video, the top-k activations per category (where $k = T_{i}/8$) of the classification activation $C(i)$ are selected and then are averaged to obtain a class-specific encoding $r_{i} \in \mathcal{R}^{C+1}$ as in ~\cite{paul2018w,Narayan_2019_ICCV}. 
	We average all the frame label predictions in the video $v_{i}$ to get the video-level ground-truth $q_{i} \in \mathcal{R}^{N_{c}+1}$.
	The video-level loss is calculated by
	
	\begin{equation}
	\mathcal{L}_{video} = -\frac{1}{N} \sum_{i=1}^{N} \sum_{j=1}^{N_{c}}   q_{i}(j)\log\frac{\exp(r_{i}(j))}{\sum_{N_{c}+1}\exp(r_{i}(k))},
	\end{equation}
	where $q_{i}(j)$ is the $j^{th}$ value of $q_{i}$ representing the probability mass of video $v_{i}$ belong to $j^{th}$ class.
	
	Consequently, the total training objective for our proposed method is
	\begin{equation}
	\mathcal{L} = \mathcal{L}_{frame} + \alpha \mathcal{L}_{video} + \beta \mathcal{L}_{actionness},
	\end{equation}
	where $\mathcal{L}_{frame}$, $\mathcal{L}_{video}, $ and $ \mathcal{L}_{actionness},$ denote the frame classification loss, video classification loss, and actionness loss, respectively. $\alpha \text{ and }\beta$ are the hyper-parameters leveraging different losses.

	\subsection{Inference} 
	
	During the test stage, we need to give the temporal boundary for each detected action.
	We follow previous weakly-supervised work \cite{nguyen2018weakly} to predict video-level labels by temporally pooling and thresholding on the classification score.
	As shown in Fig.~\ref{fig:fig2}, we first obtain the classification score $C$ and actionness score $A$ by feeding input features of a video to the classification module and actionness module.
	Towards segment localization, we follow the thresholding strategy in~\cite{nguyen2018weakly,Nguyen_2019_ICCV} to keep the action frames above the threshold and consecutive action frames constitute an action segment. 
	For each predicted video-level action class, we localize each action segment by detecting an interval that the sum of classification score and actionness score exceeds the preset threshold at every frame inside the interval.
	We simply set the confidence score of the detected segment to the sum of its highest frame classification score and the actionness score.
	Towards single frame localization, for the action instance, we choose the frame with the maximum activation score in the detected segment as the localized action frame. 
	

	\section{Experiment}
	
	\subsection{Datasets}

	\textbf{THUMOS14.}
	There are 1010 validation and 1574 test videos from 101 action categories in THUMOS14~\cite{idrees2017thumos}. Out of these, 20 categories have temporal annotations in 200 validation and 213 test videos. The dataset is challenging, as it contains an average of 15 activity instances per video. Similar to [14, 16], we use the validation set for training and test set for evaluating our framework.

	
	\noindent\textbf{GTEA.}
	There are 28 videos of 7 fine-grained types of daily activities in a kitchen contained in GTEA~\cite{lei2018temporal}. An activity is performed by four different subjects, and each video contains about 1800 RGB frames, showing a sequence of 7 actions, including the background action. 
	
	\noindent\textbf{BEOID.}
	There are 58 videos in BEOID~\cite{damen2014you}. There is an average of 12.5 action instances per video. The average length is about 60s, and there are 30 action classes in total. 
	We randomly split the untrimmed videos in an 80-20\% proportion for training and testing, as described in~\cite{moltisanti2019action}.

	\subsection{Implementation Details}
	We use I3D network~\cite{CarreiraI3D} trained on the Kinetics~\cite{carreira2017quo} to extract video features. For the RGB stream, we rescale the smallest dimension of a frame to 256 and perform the center crop of size $224\times 224$. For the flow stream, we apply the TV-L1 optical flow algorithm~\cite{zach2007duality}. 
	We follow the two-stream fusion operation in~\cite{Narayan_2019_ICCV} to integrate predictions from both appearance~(RGB) and motion~(Flow) branches. 
	The inputs to the I3D models are stacks of 16 frames.
	
	On all datasets, we set the learning rate to $10^{-3}$ for all experiments, and the model is trained with a batch size of 32 using the Adam~\cite{kingma2014adam}.
	Loss weight hyper-parameters $\alpha$ and $\beta$ are set to 1.
	The model performance is not sensitive to these hyper-parameters. 
	For the hyper-parameter $\eta$ used in mining background frames, we set it to 5 on THUMOS14 and set it to 1 on the other two datasets. The number of iterations is set to 500, 2000 and 5000 for GTEA, BEOID and THUMOS14, respectively.

	\subsection{Evaluation Metrics}
	
	(1) \textbf{Segment localization}: We follow the standard protocol, provided with the three datasets, for evaluation.
	The evaluation protocol is based on mean Average Precision (mAP) for different intersection over union~(IoU) values for the action localization task.
	
	\noindent(2) \textbf{Single-frame localization}: We also use mAP to compare performances. Instead of measuring IoU, the predicted single-frame is regarded as correct when it lies in the temporal area of the ground-truth segment, and the class label is correct. We use mAP@hit to denote the mean average precision of selected action frame falling in the correct action segment.

	\subsection{Annotation Analysis}
	
	

	\begin{table}[!t]
		\centering
		\small
		\caption{Comparison between different methods for simulating single-frame supervision on THUMOS14. ``Annotation'' means that the model uses human annotated frame for training. 
			``TS'' denotes that the single-frame is sampled from action instances using a uniform distribution, while ``TS in GT'' is using a Gaussian distribution near the mid timestamp of each activity. The AVG for segment localization is the average mAP from IoU 0.1 to 0.7. 
		}
		
		\setlength\extrarowheight{1pt}
		\begin{tabular}{l|c|p{1.7cm}p{1.7cm}p{1.7cm}p{1.7cm}}
			\hline
			\multirow{2}{*}{Position} &  \multirow{2}{*}{mAP@hit} & \multicolumn{4}{c}{Segment mAP@IoU }          \\ \cline{3-6}
			\multicolumn{1}{c|}{}  & &  0.3  & 0.5  & 0.7  & AVG    \\ \hline
			Annotation  & \textbf{60.2}$\pm$0.70  & \textbf{53.3}$\pm$0.30 & 28.8$\pm$0.57 & 9.7$\pm$0.35 & \textbf{40.6}$\pm$0.40\\
			TS & 57.6$\pm$0.60 & 52.0$\pm$0.35 & \textbf{30.2}$\pm$0.48 & \textbf{11.8}$\pm$0.35 & 40.5$\pm$0.28 \\
			TS in GT & 52.8$\pm$0.85	& 47.4$\pm$0.72	& 26.2$\pm$0.64	& 9.1$\pm$0.41 &	36.7$\pm$0.52  \\ \hline
		\end{tabular}
		\label{tab:th14_FP}
	\end{table}
	
	\subsubsection{Single-frame supervision simulation.}

	First, to simulate the single-frame supervision based on ground-truth boundary annotations existed in the above three datasets, we explore the different strategies to sample a single-frame for each action instance.
	We follow the strategy in~\cite{moltisanti2019action} to generate single-frame annotations with uniform and Gaussian distribution~(\textbf{Denoted by TS and TS in GT}).
	We report the segment localization at different IoU thresholds and frame localization results on THUMOS14 in Table~\ref{tab:th14_FP}.
	The model with each single-frame annotation is trained five times. The mean and standard deviation of mAP is reported in the Table.
	Compared to models trained on sampled frames, the model trained on human annotated frames achieves the highest mAP@hit.
	As the the action frame is the frame with the largest prediction score in the prediction segment, the model with higher mAP@hit can assist with localizing action timestamp more accurately when people need to retrieve the frame of target actions.
	When sampling frames are from near middle timestamps to the action segment~(TS in GT), the model performs inferior to other models as these frames may not contain informative elements of complete actions.
	For the segment localization result, the model trained on truly single-frame annotations achieves higher mAP at small IoU thresholds, and the model trained on frames sampled uniformly from the action instance gets higher mAP at larger IoU thresholds.	
	It may be originated by sampled frames of uniform distribution containing more boundary information for the given action instances.


	\subsubsection{Single-frame annotation.}
	
	\begin{table}[t]
		\centering
		\caption{Single-frame annotation differences between different annotators on three datasets.
			We show the number of action segments annotated by Annotator 1, Annotator 2, Annotator 3, and Annotator 4.
			In the last column, we report the total number of the ground-truth action segments for each dataset.}
		\begin{tabular}{c|cccc|c}
			\hline
			Datasets  & Annotator 1 & Annotator 2  & Annotator 3 & Annotator 4 & $\#$ of total segments \\ \hline
			GTEA     &   369 & 366 & 377 & 367 & 367 \\ 
			BEOID     &   604 & 602 & 589 & 599 & 594 \\ 
			THUMOS14   &    3014 & 2920 & 2980 & 2986 & 3007 \\ \hline
		\end{tabular}
		\label{tab:ann_num}
	\end{table}

	We also invite four annotators with different backgrounds to label a single frame for each action segment on three datasets. More details of annotation process can be found in the supplementary material. In Table~\ref{tab:ann_num}, we have shown the action instances of different datasets annotated by different annotators. The ground-truth in the Table denotes the action instances annotated in the fully-supervised setting. From the Table, we obtain that the number of action instances by different annotators have a very low variance. The number of labeled frames is very close to the number of action segments in the fully-supervised setting. This indicates that annotators have common justification for the target actions and hardly miss the action instance despite that they only pause once to annotate single-frame of each action.

	\begin{figure}[t]
		\centering
		\includegraphics[width=0.8\linewidth]{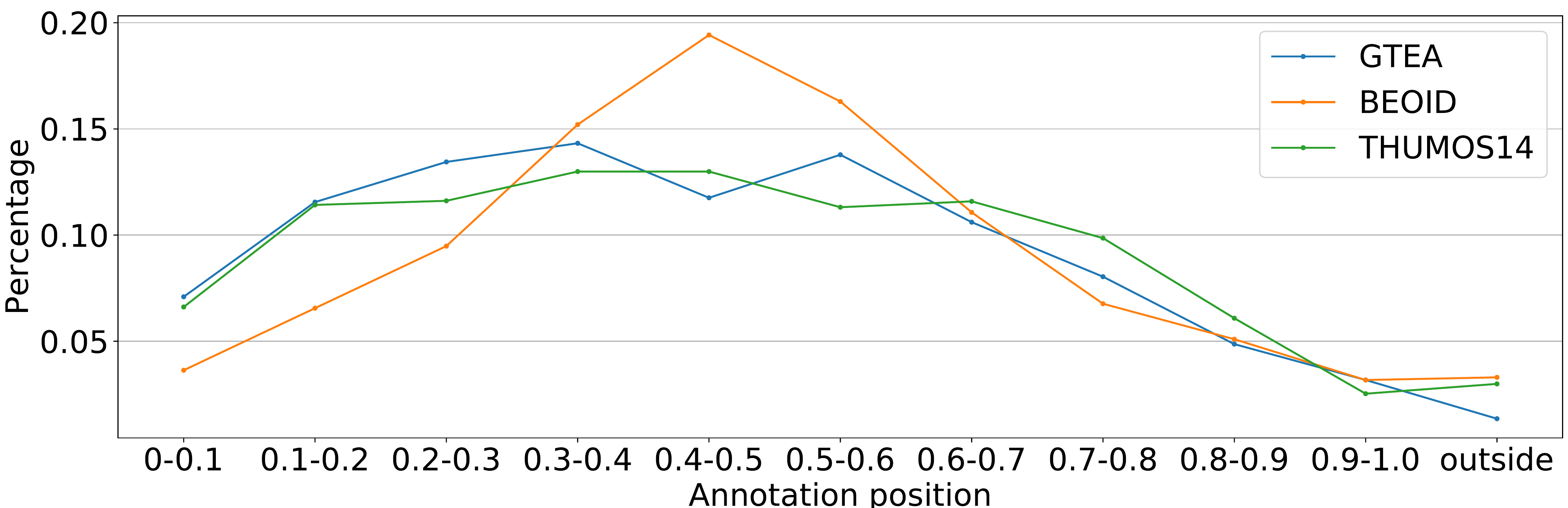}
		\caption{Statistics of human annotated single-frame on three datasets. X-axis: single-frame falls in the relative portion of the whole action; Y-axis: percentage of annotated frames. We use different colors to denote annotation distribution on different datasets. 
		}
		\label{fig:ann_distribution}
	\end{figure}
	
	We also present the distribution of the relative position of single-frame annotation to the corresponding action segment.
	As shown in Fig.~\ref{fig:ann_distribution},  there are rare frames outside of the temporal range of action instances from the ground-truth in the fully-supervised setting. 
	As the number of annotated single frames is almost the same as the number of action segments, we can draw the inference that the single frame annotation includes all almost potential action instances.  
	We obtain that annotators prefer to label frames near to the middle part of action instances. 
	This indicates that humans can identify an action without watching the whole action segment. On the other hand, this will significantly reduce the annotation time compared with fully-supervised annotation as we can quickly skip the current action instance after single-frame annotation.

	\subsubsection{Annotation speed for different supervision.}
	To measure the required annotation resource for different supervision, we conduct a study on GTEA. Four annotators are trained to be familiar with action classes in GTEA. We ask the annotator to indicate the video-level label, single-frame label and temporal boundary label of 21 videos lasting 93 minutes long. While watching, the annotator is able to skim quickly, pause, and go to any timestamp. On average, the annotation time used by each person to annotate 1-minute video is 45s for the video-level label, 50s for the single-frame label, and 300s for the segment label.
	The annotation time for single-frame label is close to the annotation time for video-level label but much fewer than time for the fully-supervised annotation.

	\begin{table}[!ht]
		\centering
		\caption{Segment localization mAP results at different IoU thresholds on three datasets. Weak denotes that only video-level labels are used for training. All action frames are used in the full supervision approach. SF uses extra single frame supervision with frame level classification loss. SFB means that pseudo background frames are added into the training, while the SFBA adopts the actionness module, and the SFBAE indicates the action frame mining strategy added in the model. For models trained on single-frame annotations, we report mean and standard deviation results of five runs.  AVG is the average mAP from IoU 0.1 to 0.7.}
		
		\setlength\extrarowheight{1pt}
		\begin{tabular}{c|l|p{1.7cm}p{1.7cm}p{1.7cm}p{1.7cm}p{1.7cm}}
			\hline
			\multicolumn{1}{c|}{\multirow{2}{*}{Dataset}}  &\multicolumn{1}{c|}{\multirow{2}{*}{Models}}     & \multicolumn{5}{c}{mAP@IoU}      \\ \cline{3-7}
			\multicolumn{1}{c|}{} & \multicolumn{1}{c|}{}         &  0.1  & 0.3 & 0.5 & 0.7 & AVG \\   \hline \hline
			\multirow{6}{*}{GTEA} 
			&Full & 58.1 & 40.0 & 22.2 & 14.8 & 31.5 \\ 
			&Weak  & 14.0 & 9.7 & 4.0 & 3.4 & 7.0 \\ \cline{2-7}
			&SF & 50.0$\pm$1.42 & 35.6$\pm$2.61 & \textbf{21.6}$\pm$1.67 & \textbf{17.7}$\pm$0.96 & 30.5$\pm$1.23 \\ 
			&SFB & 52.9$\pm$3.84 & 34.9$\pm$4.72 & 17.2$\pm$3.46 & 11.0$\pm$2.52 & 28.0$\pm$3.53 \\ 
			&SFBA & 52.6$\pm$5.32 & 32.7$\pm$3.07 & 15.3$\pm$3.63 & 8.5$\pm$1.95 & 26.4$\pm$3.61 \\ 
			&SFBAE & \textbf{58.0}$\pm$2.83 & \textbf{37.9}$\pm$3.18 & 19.3$\pm$1.03 & 11.9$\pm$3.89 & \textbf{31.0}$\pm$1.63 \\ \hline
			\multirow{6}{*}{BEOID} 
			&Full & 65.1 & 38.6 & 22.9 & 7.9 & 33.6 \\ 
			&Weak & 22.5 & 11.8 & 1.4 & 0.3 & 8.7 \\ \cline{2-7}
			&SF & 54.1$\pm$2.48 & 24.1$\pm$2.37 & 6.7$\pm$1.72 & 1.5$\pm$0.84 & 19.7$\pm$1.25 \\ 
			&SFB & 57.2$\pm$3.21 & 26.8$\pm$1.77 & 9.3$\pm$1.94 & 1.7$\pm$0.68 & 21.7$\pm$1.43 \\ 
			&SFBA & \textbf{62.9}$\pm$1.68 & 36.1$\pm$3.17 & 12.2$\pm$3.15 & 2.2$\pm$2.07 & 27.1$\pm$1.44 \\ 
			&SFBAE & \textbf{62.9}$\pm$1.39 & \textbf{40.6}$\pm$1.8 & \textbf{16.7}$\pm$3.56 & \textbf{3.5}$\pm$0.25 & \textbf{30.1}$\pm$1.22 \\ \hline

			\multirow{6}{*}{THUMOS14} 
			
			&Full & 68.7 & 54.5 & 34.4 & 16.7 & 43.8 \\ 
			&Weak & 55.3 & 40.4 & 20.4 & 7.3 & 30.8 \\ \cline{2-7}
			&SF & 58.6$\pm$0.56 & 41.3$\pm$0.62 & 20.4$\pm$0.55 & 6.9$\pm$0.33 & 31.7$\pm$0.41 \\ 
			&SFB & 60.8$\pm$0.65 & 44.5$\pm$0.37 & 22.9$\pm$0.38 & 7.8$\pm$0.46 & 33.9$\pm$0.31 \\ 
			&SFBA & 68.7$\pm$0.33 & 52.3$\pm$1.21 & 28.2$\pm$0.42 & \textbf{9.7}$\pm$0.51 & 39.9$\pm$0.43 \\ 
			&SFBAE & \textbf{70.0}$\pm$0.64 & \textbf{53.3}$\pm$0.3 & \textbf{28.8}$\pm$0.57 & \textbf{9.7}$\pm$0.35 & \textbf{40.6}$\pm$0.40 \\ \hline

		\end{tabular}
		\label{tab:ab}
	\end{table}

	\subsection{Analysis}
	
	\subsubsection{Effectiveness of each module, loss, and supervision.}
	
	To analyze the contribution of the classification module, actionness module, background frame mining strategy, and the action frame mining strategy, we perform a set of ablation studies on THUMOS14, GTEA and BEOID datasets. The segment localization mAP at different thresholds is presented in Table~\ref{tab:ab}. We also compare the model with only weak supervision and the model with full supervision. The model with weak supervision is implemented based on~\cite{Narayan_2019_ICCV}.
	
	We observe that the model with single-frame supervision outperforms the weakly-supervised model. And large performance gain is obtained on GTEA and BEOID datasets as the single video often contains multiple action classes, while action classes in one video are fewer in THUMOS14. Both background frame mining strategy and action frame mining strategy boost the performance on BEOID and THUMOS14 by putting more frames into the training, the performance on GTEA decreases mainly due to that GTEA contains almost no background frame. In this case, it is not helpful to employ background mining and the actionness module which aims for distinguishing background against action. The actionness module works well for the BEOID and THUMOS14 datasets, although the actionness module only produces one score for each frame.
	

	\begin{table*}[!t]
		\centering
		\setlength\extrarowheight{1pt}
		\caption{Segment localization results on THUMOS14 dataset. The mAP values at different IoU thresholds are reported, and the column AVG indicates the average mAP at IoU thresholds from 0.1 to 0.5. * denotes the single-frame labels are simulated based on the ground-truth annotations. $^\#$ denotes single-frame labels are manually annotated by human annotators.}
		\resizebox{\linewidth}{!}{
			\begin{tabular}{c|l|p{0.8cm}p{0.8cm}p{0.8cm}p{0.8cm}p{0.8cm}p{0.8cm}p{0.8cm}p{0.8cm}}
				\hline
				\multicolumn{1}{c|}{\multirow{2}{*}{Supervision}} & \multicolumn{1}{c|}{\multirow{2}{*}{Method}} & \multicolumn{8}{c}{mAP @IoU}                                  \\ \cline{3-10}
				\multicolumn{1}{c|}{}                             &                         & 0.1  & 0.2  & 0.3  & 0.4  & 0.5  & 0.6  & 0.7  & AVG \\ \hline
				Full                                             & S-CNN~\cite{Shou_2016_CVPR}                   & 47.7 & 43.5 & 36.3 & 28.7 & 19.0 & -    & 5.3  & 35.0         \\ 
				Full                                             & CDC~\cite{Shou_2017_CVPR}                   & - & - & 40.1 & 29.4 & 23.3 & -    & 7.9  & -         \\ 
				Full                                             & R-C3D~\cite{xu2017r}                   & 54.5 & 51.5 & 44.8 & 35.6 & 28.9 & -    & -    & 43.1         \\
				Full                                             & SSN~\cite{zhao2017temporal}                     & 60.3 & 56.2 & 50.6 & 40.8 & 29.1 & -    & -    & 47.4         \\
				Full                                             & Faster-~\cite{chao2018rethinking}                 & 59.8 & 57.1 & 53.2 & 48.5 & 42.8 & \textbf{33.8} & \textbf{20.8} & 52.3         \\
				Full & BMN~\cite{Lin_2019_ICCV} & - & - & 56.0 & 47.4 & 38.8 & 29.7 & 20.5 & -\\
				Full & P-GCN~\cite{Zeng_2019_ICCV} & \textbf{69.5} & \textbf{67.8} &\textbf{63.6} &\textbf{57.8} &\textbf{49.1} & - & - & \textbf{61.6}\\ \hline
				Weak                                             & Hide-and-Seek~\cite{singh2017hide}           & 36.4 & 27.8 & 19.5 & 12.7 & 6.8  & -    & -    & 20.6         \\
				Weak                                             & UntrimmedNet~\cite{wang2017untrimmednets}            & 44.4 & 37.7 & 28.2 & 21.1 & 13.7 & -    & -    & 29.0         \\
				Weak                                             & W-TALC~\cite{ding2018weakly}                  & 49.0 & 42.8 & 32.0 & 26.0 & 18.8 & -    & 6.2  & 33.7         \\
				Weak                                             & AutoLoc~\cite{shou2018autoloc}                 & -    & -    & 35.8 & 29.0 & 21.2 & 13.4 & 5.8  & -            \\
				Weak & STPN~\cite{nguyen2018weakly}                  &  52.0    &  44.7   &     35.5  &  25.8  & 16.9  & 9.9    & 4.3   &  35.0 \\
				Weak & W-TALC~\cite{paul2018w} & 55.2 & 49.6 & 40.1 & 31.1 & 22.8 & - & 7.6 & 39.7 \\
				Weak & Liu~\etal~\cite{Liu_2019_CVPR}                 & 57.4    & 50.8    & 41.2 & 32.1 & 23.1 & 15.0 & 7.0  & 40.9            \\
				Weak                                             & Nguyen~\etal ~\cite{Nguyen_2019_ICCV}                 & \textbf{60.4}    & \textbf{56.0}    & \textbf{46.6} & \textbf{37.5} & \textbf{26.8} & \textbf{17.6} & \textbf{9.0}  & \textbf{45.5}            \\
				Weak                                             & 3C-Net~\cite{Narayan_2019_ICCV}                  & 59.1 & 53.5 & 44.2 & 34.1 & 26.6 & -    & 8.1  &  43.5            \\ \hline
				Single-frame simulation*											& Moltisanti~\etal~\cite{moltisanti2019action} 			&24.3 & 19.9 & 15.9 & 12.5 & 9.0 & -        & -      & 16.3     \\ 
				Single-frame simulation*                                            & SF-Net                    & 68.3 & 62.3 & 52.8 & \textbf{42.2} & \textbf{30.5} & \textbf{20.6} & \textbf{12.0} & 51.2    \\ 
				\hline
				Single-frame$^\#$                                            & SF-Net                    & \textbf{71.0}  & \textbf{63.4} & \textbf{53.2} & 40.7 & 29.3 & 18.4 & 9.6 & \textbf{51.5} \\\hline  
		\end{tabular}}
		
		\label{tab:th14}
	\end{table*} 
	

	


	\subsubsection{Comparisons with state-of-the-art.}
	
	Experimental results on THUMOS14 testing set are shown in Table~\ref{tab:th14}. 
	Our proposed single-frame action localization method is compared to existing methods for weakly-supervised temporal action localization, as well as several fully-supervised ones. 
	Our model outperforms the previous weakly-supervised methods at all IoU thresholds regardless of the choice of feature extraction network. The gain is substantial even though only one single-frame for each action instance is provided.
	The model trained on human annotated frames achieves higher mAP at lower IoU compared to model trained on sampling frames uniformly from action segments. The differences come from the fact that the uniform sampling frames from ground-truth action segments contain more information about temporal boundaries for different actions.
	As there are many background frames in the THUMOS14 dataset, the single frame supervision assists the proposed model with localizing potential action frames among the whole video.
	Note that the supervised methods have the regression module to refine the action boundary, while we simply threshold on the score sequence and still achieve comparable results.

	\section{Conclusions}
	In this paper, we have investigated how to leverage single-frame supervision to train temporal action localization models for both segment localization and single-frame localization during inference.
	Our SF-Net makes full use of single-frame supervision by predicting actionness score, pseudo background frame mining and pseudo action frame mining. 
	SF-Net significantly outperforms weakly-supervised methods in terms of both segment localization and single-frame localization on three standard benchmarks. 
	
	\noindent\textbf{Acknowledgements}.
	The authors from UTS were partially supported by ARC DP200100938 and Facebook.

	\clearpage
	%
	%
	\bibliographystyle{splncs04}
	\bibliography{reference}
	
	\clearpage
	\appendix
	\noindent\textbf{\Large Appendix}

	\section{Single-frame Annotation}
	We invite four annotators with different backgrounds to label single-frames for all actions intances. Before annotating each dataset, four annotators have watched a few video examples containing different actions to be familiar with action classes.
	They are asked to annotate one single frame for each target action instance while watching the video by our designed annotation tool.
	Specifically, they are required to pause the video when they identify an action instance and choose the action class that the paused frame belongs to. Once they have chosen the action class, they need to continue watching the video and record the frames for the next target action instances.
	After watching the whole video, the annotator should press the generation button and the annotation tool will then automatically produce the timestamps and action classes of all operated frames for the given video.
	Compared to the annotation process in the weakly-supervised setting, this results into almost no extra time cost since the timestamps are automatically generated.
	The single-frame annotation process is much faster than annotating the temporal boundary of each action in which the annotator often watches the video many times to define the start and end timestamp of a given action. 
	\subsection{Annotation guideline}
	Different people may have different understandings of what constitutes a given action. To reduce the ambiguity, we prepare a detailed annotation guideline, which includes both clear action definitions as well as positive/negative examples with detailed clarifications for each action. 
	For each action, we give (1) textual action definition for single-frame annotation, (2) positive single-frame annotations, and (3) segmented action instances for annotator to be familiar with.
	
	\subsection{Annotation tool}
\begin{figure}[!t]
		\begin{center}
			\includegraphics[width=\linewidth]{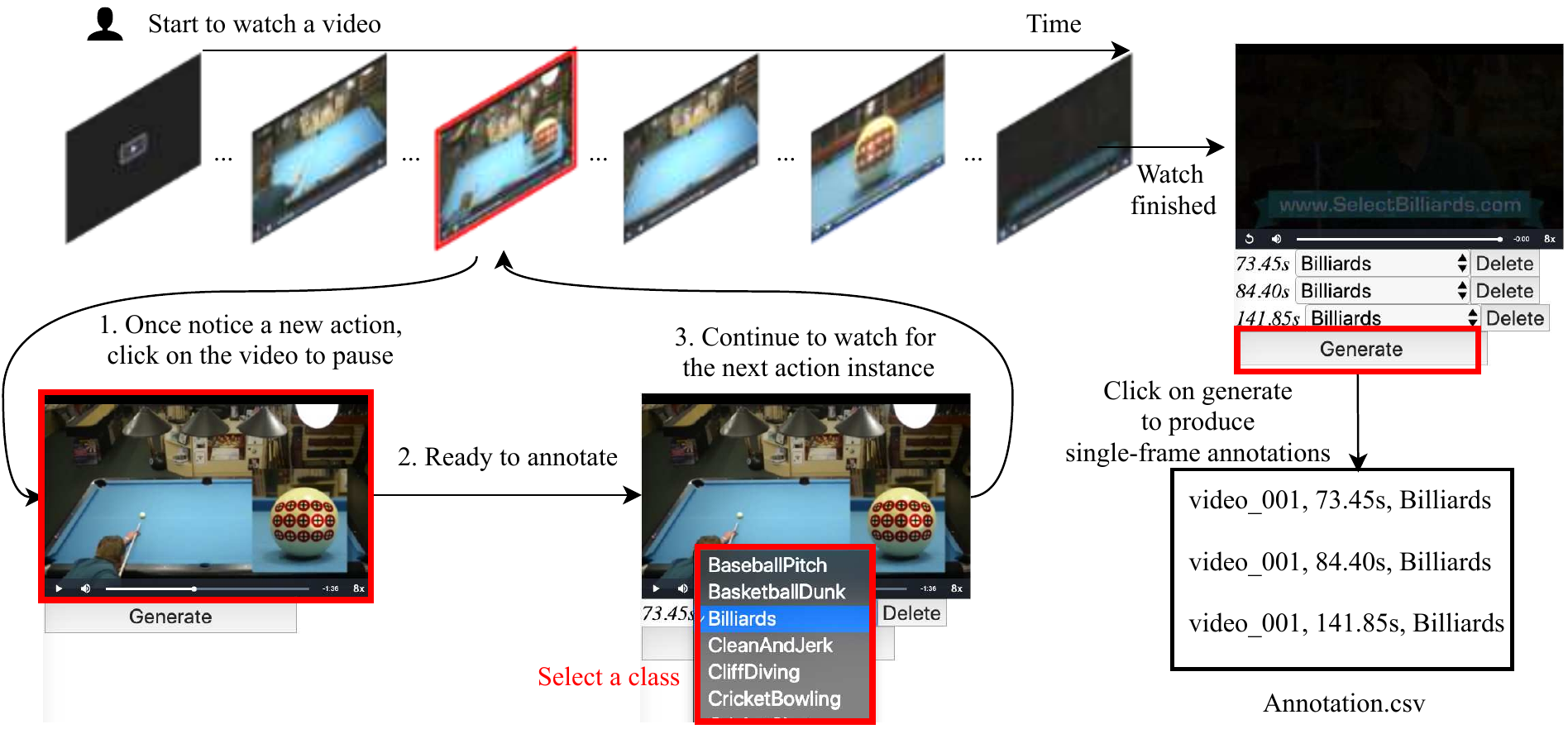}
		\end{center}
		\caption{Interface for annotating a single frame. First step is to pause the video when annotators notice an action while watching the video. The second step is to select the target class for the paused frame. After annotating an action instance, the annotator can click the video to keep watching the video for the next action instance. Note that the time is automatically generated by the annotation tool. After watching a whole video, the annotator can press the generate button to save all records into a csv file.  }
		\label{fig:ann}
	\end{figure}

	Our annotation tool supports automatically recording timestamp for annotating single-frame. This makes the annotation process faster when annotators notice an action and ready to label the paused frame. The interface of our annotation tool is presented in Figure~\ref{fig:ann}. After watching a whole video, the annotator can press the generate button, the annotation results will be automatically saved into a csv file. When annotators think they made a wrong annotation, they can delete it at any time while watching the video. We have shown the one annotation example in the supplementary file. We have uploaded a video in the supplementary file to show how to annotate single-frame while watching the video.
	
	\subsection{Quality control}
	We make two efforts to improve the annotation quality. First of all, each video is labeled by four annotators, and the annotated single-frames of a view are randomly selected during experiments to reduce annotation bias. Secondly, we train annotators before annotating videos and make sure that they can notice target actions while watching the video.
	
	\begin{algorithm}[]
		
		\begin{algorithmic}[1]
			\caption{Action Frame Mining}\label{alg1}
			
			\State{\bfseries Input:}  video classification activation $C\in \mathcal{R}^{T\times N_{c}+1}$, labeled action frame at time t belonging to action class $y$, expand radius $r=5$, threshold $\xi=0.9$. 
			\State{\bfseries Output:} expanded frames set $\mathcal{S}$ 
			\State{gather classification score  $C(t)$ for the anchor frame}
			\State{$\mathcal{S} \leftarrow \{(t, y)\} $}
			\Function{Expand}{s} ;
			\For{$j\leftarrow1;j\leq r$} 
			\State{$\hat{y}_{past} \leftarrow  \text{argmin} ~C(t + (j-1)s)$}
			\State{$\hat{y}_{current} \leftarrow \text{argmin}~ C(t + js)$}
			\State{$\hat{y}_{future} \leftarrow \text{argmin} ~C(t + (j+1)s)$}
			\If{$\hat{y}_{past} == \hat{y}_{current} == \hat{y}_{future}$ and $ C(t + js)_{y}\geq \xi C(t)_{y}$}
			\State{$\mathcal{S} \leftarrow (t + js, y)$}
			\EndIf
			\State{$j\leftarrow j+s$}
			\EndFor
			
			\EndFunction
			\State{EXPAND(-1)}
			\State{EXPAND(1)}	
			\State{Return $\mathcal{S}$}
		\end{algorithmic}
	\end{algorithm}

	\section{Action Frame Mining}    	
	The action frame mining strategy is described in Algorithm~\ref{alg1}.
	We treat the labeled action frame as the anchor frame and expand frames around it.
	We use a threshold $\xi$ and the label consistency with neighbors to decide whether to add the unlabeled frame or not.
	The expanded frames are annotated with the same label as the anchor frame.  
	As shown in Algorithm~\ref{alg1}, we expand the frames at $t-1$ to the anchor frame.
	We first gather the classification score of three frames around $t-1$ frame. We then calculate the prediction classes for these three frames. If they all have the same predicted class and the classification score for the current frame at class $y$ is above a threshold, we choose to put the current frame into the training frame set $S$. For all experiments in the current paper, we set $\xi=0.9$ for fair comparison.
	
	\section{Evaluate Classification \& Localization Independently}
	We evaluate our single-frame supervised model and weakly-supervised model in terms of classification and localization independently.
	We adopt mean average precision~(mAP) in~\cite{TSN2016ECCV} to evaluate the video-level classification performance and AP at different IoU thresholds to evaluate the class-agnostic localization quality regardless of the action class.
	We report the video-level classification mAP in Table~\ref{tab:act}, showing only marginal gain as expected. 
	This is because THUMOS14 only contains one or two action classes in a single video which makes the video be easily classified into the target action category.
	We also evaluates boundary detection AP regardless of the label in Table~\ref{tab:act}, showing large gain after adding single-frame supervision.
	
	\begin{table}[!t]
		\centering
		
		\caption{Classification accuracy and class-agnostic localization AP on THUMOS14.}
		\begin{tabular}{l|c|ccc}
			\hline
			& Classification & \multicolumn{3}{c}{Class-agnostic localization} \\ \cline{2-5}
			&  mAP   & AP@IoU=0.3 & AP@IoU=0.5 & AP@IoU=0.7\\ \hline
			Ours w/o single-frame & 97.8 &  42.1  & 18.1 & 5.5\\
			Ours w/ single-frame &   \textbf{98.5} & \textbf{58.8} & \textbf{32.4} & \textbf{9.4}\\
			\hline
		\end{tabular}
		\label{tab:act}
	\end{table}

	\begin{table}[!t]
		\centering
		\small
		\setlength\extrarowheight{2pt}
		\caption{The background $\eta$ analysis on THUMOS14. AVG is the average mAP at IoU 0.1 to 0.7.}
		\begin{tabular}{>{\centering} m{0.8cm}|c|cccccc}
			\hline
			\multicolumn{1}{c|}{\multirow{2}{*}{$\eta$}} & \multirow{2}{*}{mAP@hit} &\multicolumn{6}{c}{mAP@IoU }                             \\ \cline{3-8}
			\multicolumn{1}{c|}{}   & & 0.1  & 0.3 & 0.5 & 0.6  & 0.7  & AVG     \\ \hline
			0.0 & 44.4$\pm$0.56 & 58.6$\pm$0.55 & 41.1$\pm$0.80 & 20.2$\pm$0.69 & 12.9$\pm$0.58 & 7.3$\pm$0.10 & 31.7$\pm$0.47 \\
			1.0 & 57.7$\pm$0.41 & 68.3$\pm$0.37 & 51.1$\pm$0.57 & 28.2$\pm$0.52 & 17.7$\pm$0.09 & 9.4$\pm$0.31 & 39.3$\pm$0.13 \\ 
			3.0 & 60.6$\pm$1.36 & 71.0$\pm$1.21 & 53.8$\pm$0.71 & 29.3$\pm$1.14 & 18.9$\pm$0.88 & 9.4$\pm$0.43 & 41.1$\pm$0.80 \\ 
			5.0 & 60.6$\pm$0.85 & 70.6$\pm$0.92 & 53.7$\pm$1.21 & 29.1$\pm$0.39 & 19.1$\pm$1.31 & 10.2$\pm$0.84 & 41.1$\pm$0.78 \\ 
			7.0 & 60.9$\pm$0.56 & 70.7$\pm$0.08 & 54.3$\pm$1.18 & 29.5$\pm$0.13 & 19.0$\pm$0.50 & 10.1$\pm$0.27 & 41.3$\pm$0.44 \\ 
			9.0 & 60.2$\pm$1.12 & 70.3$\pm$0.83 & 53.4$\pm$0.8 & 29.6$\pm$0.58 & 18.8$\pm$0.99 & 10.1$\pm$0.37 & 41.0$\pm$0.60 \\ \hline
			
		\end{tabular}
		\label{tab:eta}
	\end{table}

	\begin{table}[!t]
		\centering
		\small
		\setlength\extrarowheight{2pt}
		\caption{The loss coefficients analysis on THUMOS14. AVG is the average mAP at IoU 0.1 to 0.7.}
		\begin{tabular}{l|c|ccccccc}
			\hline
			\multirow{2}{*}{parameter} & \multirow{2}{*}{mAP@hit} & \multicolumn{6}{c}{Segment mAP@IoU} \\ \cline{3-8}
			\multicolumn{1}{c|}{}   & & 0.1 & 0.3 & 0.5 & 0.6  & 0.7  & AVG   \\ \hline
			$\alpha=0.2$ & 61.9$\pm$0.34 & 71.6$\pm$0.73 & 54.2$\pm$1.31 & 29.3$\pm$0.47 & 18.4$\pm$0.62 & 9.7$\pm$0.35 & 41.3$\pm$0.56 \\ 
			$\alpha=0.5$ & 61.9$\pm$0.68 & 71.8$\pm$0.36 & 54.4$\pm$0.68 & 30.2$\pm$0.41 & 19.3$\pm$0.92 & 10.2$\pm$1.14 & 41.9$\pm$0.47 \\ 
			$\alpha=0.8$ & 60.7$\pm$0.95 & 71.0$\pm$0.40 & 53.8$\pm$0.64 & 29.4$\pm$0.26 & 19.0$\pm$0.23 & 10.0$\pm$0.25 & 41.2$\pm$0.22 \\ \hline
			$\beta=0.2$ & 60.6$\pm$1.55 & 70.5$\pm$1.21 & 53.2$\pm$1.09 & 29.4$\pm$0.64 & 18.8$\pm$0.71 & 9.7$\pm$0.33 & 41.0$\pm$0.67 \\ 
			$\beta=0.5$ & 60.2$\pm$0.69 & 70.5$\pm$0.55 & 53.7$\pm$0.71 & 29.4$\pm$0.16 & 18.8$\pm$0.47 & 10.0$\pm$0.34 & 41.1$\pm$0.42 \\ 
			$\beta=0.8$ & 60.8$\pm$1.05 & 70.6$\pm$0.50 & 53.8$\pm$1.47 & 29.6$\pm$0.34 & 18.9$\pm$0.36 & 10.0$\pm$0.37 & 41.2$\pm$0.55 \\ \hline
			
		\end{tabular}
		\label{tab:hypers}
	\end{table}

	\section{Sensitivity Analysis}
	\subsection{Background Ratio}
	Table~\ref{tab:eta} shows the results with respect to different background ratios $\eta$ on THUMOS14. 
	The mean and standard deviation of segment and frame metrics are reported. We ran each experiment three times. The single-frame annotation for each video is randomly sampled from annotations by four annotators.
	From the table~\ref{tab:eta}, we find that our proposed SF-Net boosts the segment and frame evaluation metrics on THUMOS14 dataset with background mining. 
	The model becomes stable when the $\eta$ is set in range from 3 to 9.
	
	\subsection{Loss coefficients}
	We also conduct experiments to analyze the hyper-parameters of each loss item on the THUMOS14 in Table~\ref{tab:hypers}.
	The mean and standard deviation of segment and frame metrics are reported. We ran each experiment three times. The single-frame annotation for each video is randomly sampled from annotations by four annotators.
	The default values of $\alpha$ and $\beta$ are 1. We change one hyper-parameter and fix the other one.
	From the Table~\ref{tab:hypers}, we observe that our model is not sensitive to the hyper-parameters.

	\begin{table}[!t]
		\centering
		\footnotesize
		\setlength\extrarowheight{2pt}
		\begin{tabular}{c|l|cccc}
			\hline
			\multicolumn{1}{c|}{\multirow{2}{*}{Supervision}} & \multicolumn{1}{c|}{\multirow{2}{*}{Method}} & \multicolumn{4}{c}{mAP @IoU}                                  \\ \cline{3-6}
			\multicolumn{1}{c|}{}                               &                      & 0.5  & 0.7  & 0.9  & AVG   \\ \hline
			Full & CDC ~\cite{Shou_2017_CVPR}    &  45.3                     &					-         &   -       &      23.8   \\ 
			Full & SSN ~\cite{zhao2017temporal}    &  41.3                     &					30.4         &   13.2        &      28.3     \\ \hline
			Weak & UntrimmedNet~\cite{wang2017untrimmednets}     & 7.4      &   3.9    &    1.2       & 3.6  \\   
			Weak & AutoLoc~\cite{shou2018autoloc} &            	27.3			&		17.5			& 		6.8			& 	16.0				\\   
			Weak & W-TALC~\cite{ding2018weakly} &            	37.0		&		14.6	& 	-			& 	18.0				\\  
			Weak & Liu~\etal~\cite{Liu_2019_CVPR} &            	36.8		&		-	& 	-			& 	\textbf{22.4}				\\  
			Weak & 3C-Net~\cite{Narayan_2019_ICCV} &            	\textbf{37.2}		&		\textbf{23.7}	& 	\textbf{9.2}			& 	21.7				\\ \hline
			Single-frame & SF-Net~(\textbf{Ours}) &   		37.8	&		24.6			& 		10.3		& 		22.8		\\  \hline
		\end{tabular}
		\caption{Segment localization results on ActivityNet1.2 validation set. The AVG indicates the average mAP from IoU 0.5 to 0.95.}
		\label{tab:act12}
	\end{table}

	\section{ActivityNet Simulation Experiment}
	We conduct experiments on ActiviytNet1.2 by randomly sampling single-frame annotations from ground truth temporal boundaries. Table~\ref{tab:act12} presents the results on ActivityNet1.2 validation set. In this experiment, the annotations are generated by randomly sampling single frame from ground truth segments.
	We follow the standard evaluation protocal~\cite{caba2015activitynet} by reporting the mean mAP scores at different thresholds~(0.5:0.05:0.95).
	On the large scale dataset, our proposed method can still obtain a performance gain with single frame supervision.

	\begin{figure}[]
		\begin{center}
			\includegraphics[width=\linewidth]{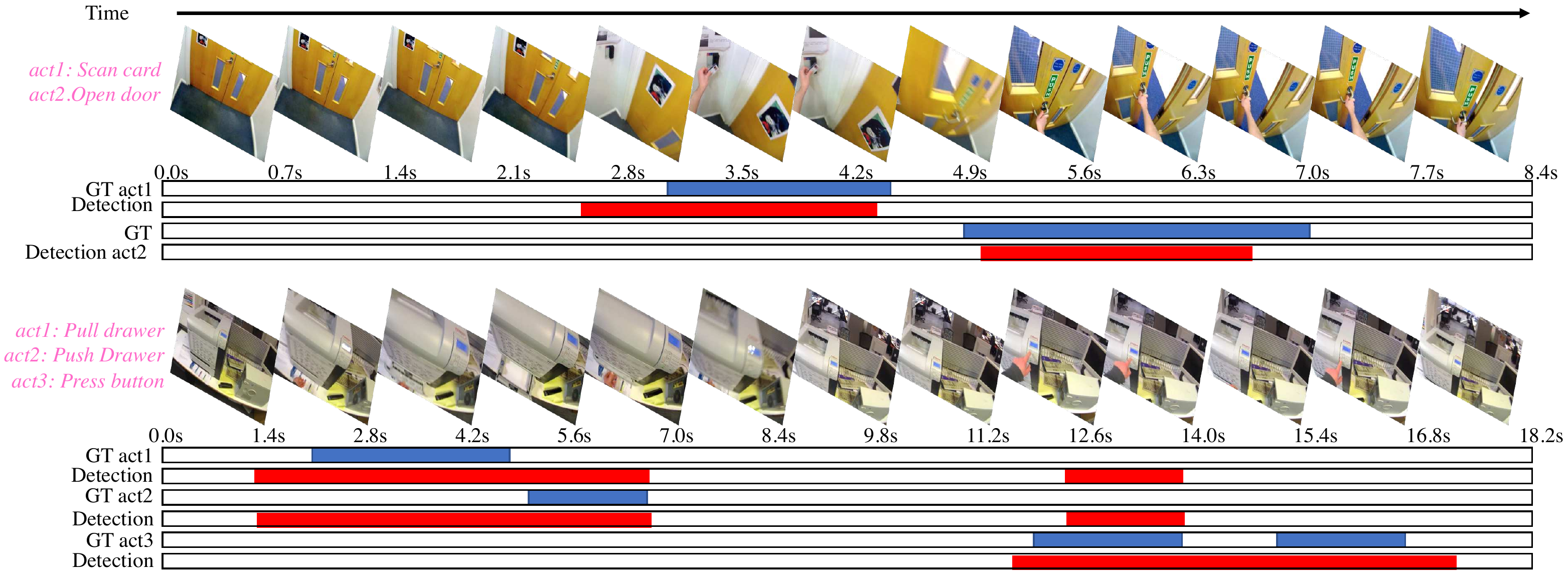}
		\end{center}
		\caption{Qualitative Results on BEOID dataset. GT denotes the ground truth and the action segment is marked with blue.
			Our proposed method detects all the action instances in the videos.}
		\label{fig:qualitative}
	\end{figure}
	
	\section{Qualitative Results}  
	
	We present the qualitative results on BEOID dataset in Figure~\ref{fig:qualitative}.
	The first example has two action instances: \textit{scan card} and  \textit{open door}.
	Our model localizes every action instance and classifies each action instance into the correct category. The temporal boundary for each instance is also close to the ground-truth annotation despite that we do not have any temporal boundary information during training.
	For the second example, there are three different actions and total four action instances.
	Our SF-Net has detected all the positive instances in the videos.
	The drawback is that the number of detected segments for each action class is greater than the number of ground truth segments. 
	To better distinguish actions of different classes,  the model should encode the fine-grained action information from the target action area instead of the 1D feature directly extracted from the whole frame. 
	We will consider this in the future work.

\end{document}